\documentclass{article}
\usepackage{spconf,amsmath,graphicx,booktabs,comment,multirow,float,cite,xcolor,microtype,nicefrac,url}
\usepackage[breaklinks,colorlinks]{hyperref}
\hypersetup{colorlinks,linkcolor={blue},citecolor={blue},urlcolor={blue}}

\usepackage[capitalize]{cleveref}

\title{Robust Localization of key fob using Channel Impulse Response of Ultra Wide Band Sensors for Keyless Entry Systems}
\name{Abhiram Kolli$^{1}$$^{2}$\sthanks{Correspondence: kolli@student.tugraz.at}\sthanks{This work was funded by the Austrian Research Promotion Agency (FFG) under the research project SEAMAL-Front (No.: 880598).  Performed the work while at Graz University of Technology, Austria.}, Filippo Casamassima$^{3}$\sthanks{Worked on dataset creation and feature extraction}, Horst Possegger$^{1}$, Horst Bischof $^{1}$}
\address{$^{1}$Graz University of Technology, Austria \hspace{.31cm}$^{2}$Profactor GmbH, Austria \hspace{.31cm}
$^{3}$NXP Semiconductors, Austria }
\begin{document}
%
\maketitle
\begin{abstract}
Using neural networks for localization of key fob within and surrounding a car as a security feature for keyless entry is fast emerging. In this paper we study: 1) the performance of pre-computed features of neural networks based UWB (ultra wide band) localization classification forming the baseline of our experiments. 2) Investigate the inherent robustness of various neural networks; therefore, we include the study of robustness of the adversarial examples without any adversarial training in this work. 3) Propose a multi-head self-supervised neural network architecture which outperforms the baseline neural networks without any adversarial training. The model's performance improved by 67\%  at certain ranges of adversarial magnitude for fast gradient sign method and 37\% each for basic iterative method and projected gradient descent method.

\end{abstract}
\section{Introduction}
\label{sec:intro}

Keyless entry, start and stop features are widely being adapted by automobile manufacturers. Estimating the location and distance information of the key fob (vehicle key) to the vehicle before executing a desired action is an additional security feature for keyless systems. Edge computing compatible neural network models are being researched to perform such tasks \cite{kolli2022test}.

The deep learning (DL) algorithms have surpassed the human level recognition performance in the image classification task \cite{langlotz2019roadmap}. Such impressive performance can be attributed to data driven models \cite{taylor2018improving}, \cite{mikolajczyk2018data} and high-performance computing. Inspired by the neural network‘s performance in image recognition task, similar concepts were adopted to other domains such as video, audio, light detection and ranging (LIDAR), etc. Contrary higher accuracy, neural networks are vulnerable towards adversarial examples (AdX) \cite{carlini2017towards}. An adversarial example is an engineered noise that is intentionally added to a clean input sample to make a well-trained deep neural network (DNN) model to misclassify. Thereby reducing the accuracy of the trained model.

\begin{figure}[t!]
  \centering
   \includegraphics[width=0.47\textwidth, height=0.21\textheight]{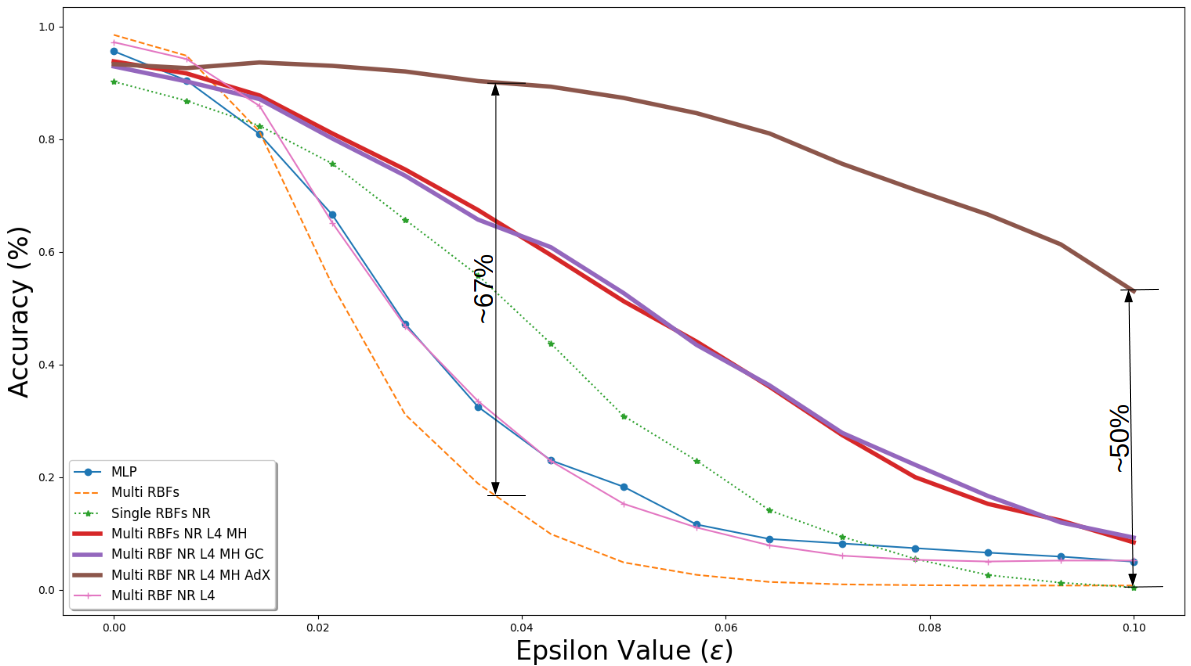}
   \caption{Adversarial robustness of our RBF-L4 neural network with multi-head model. X-axis indicates the magnitude of the adversarial noise ($\epsilon$), y-axis indicates the accuracy of the model. Ideally, higher the accuracy even at higher values of $\epsilon$ is desired. The values indicate improved performance of our model over standard architectures.}
   \label{fig:AdRgraph}
\end{figure}

Adversarial training and test-time adversarial robust methods are two types of adversarial robustness methods to overcome the adversarial attacks \cite{kolli2022test}. Typically in an adversarial training the network is trained for all possible adversarial examples. The disadvantages of such a system are firstly, the network performance degrades when classifying a clean sample. Secondly, the selection of adversarial example's magnitude which is a hyperparameter and is attacker's choice. Thus it cannot be determined at the time of training the neural network. 
To overcome such deficiencies, test-time adversarial robust methods are under active research~\cite{kolli2022test}. Though these methods outperform the traditional adversarial training, often they need to compute the gradients at the inference time for error computation. Another disadvantage of such methods is that these are iterative process and computationally expensive~\cite{kolli2022test}. This hinders the throughput of the network.

\begin{figure*}[t!]
  \centering
   \includegraphics[width=\textwidth, height=6cm]{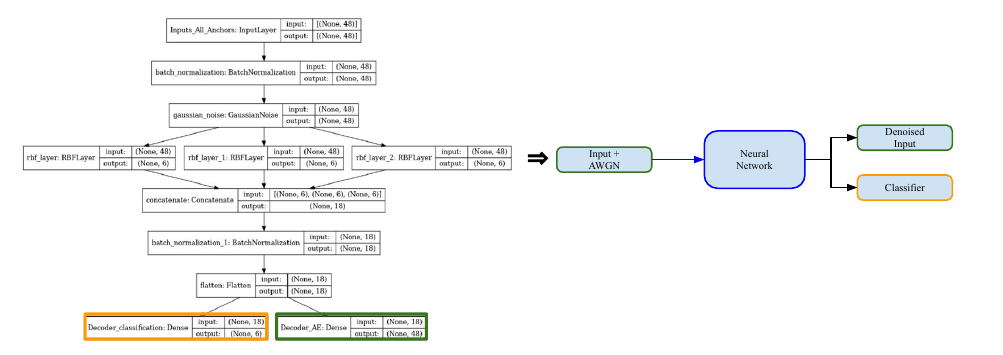}
   \caption{Architecture of self-supervised network with denoising encoder-decoder architecture (illustrated by green box) and classifier output (illustrated by orange box). The input sample is added with additive white Gaussian noise (AWGN).}
   \label{fig:selfSupgraph}
\end{figure*}

In this work, we attempt to overcome the above stated deficiencies through (see networks performance \cref{fig:AdRgraph}): firstly, we introduce a robust radial basis function (RBF) neuron dynamics with cascaded basis response (see \cref{eq:multirbf4}) which is inherently  robust to adversarial attacks (see \cref{fig:AdRgraph}). Secondly, introduced noise regularized self-supervised denoising multi-head neural architecture (see \cref{fig:selfSupgraph}) thereby extracting robust embeddings. Thirdly, designed network with smaller footprint that can be deployed on to a micro controller and also improve the throughput. \cref{fig:AdRgraph} compares our model with a baseline model. Due to lack of similar research and the specific structure of the features (see \cref{sec:data}) we considered multi-layer perceptron (MLP) as our base model. 

\section{Related Work}
\label{sec:related}
Passenger occupancy detection is one of the widely studied research that is closer to localization of the humans within and surrounding the car with a key fob. So we state the relevant known works for the passenger localization as our related work.

\subsection{Visual Sensors}
Despite growing concerns of privacy ~\cite{vamsi2020vehicle}, ~\cite{kim2019lightweight} used NIR and RGB camera for occupancy detection and people counting. To overcome the illumination constrains ~\cite{erlik2019vehicle} used the thermal cameras. ~\cite{lee2020framework} placed cameras on either side of a road to count the number of passengers present in a car/vehicle.

\subsection{Non-visual Sensors}
Non-vision systems such as capacitive sensors~\cite{george2009seat}, weight sensor~\cite{gray2002vehicle}, electromagnetic wave reflection~, Wi-Fi~\cite{xu2020wireless}, FMCW radar~\cite{song2021vehicle}, UWB senors~\cite{ma2020carosense} and ~\cite{kolli2022test}. However, these methodologies either do not consider the privacy issues or robustness of the machine learning models with the exception of ~\cite{kolli2022test}. In this work we consider the accuracy and robustness of the pre-computed features without augmentations and including adversarial examples in the training data.

\section{Methodology}
\label{sec:mtd}
The vulnerability of DNNs towards adversarial examples is due to the selection of training data and its representation, augmentation strategies, type of neurons, network architecture, loss minimization strategies, regularization strategies, etc. So, we formulated few strategies to study the inherent adversarial robustness of networks against the fast gradient sign method (FGSM), basic iterative method (BIM) and projected gradient descent (PGD) adversarial samples. These attacks are modeled as mentioned in ~\cite{ren2020adversarial}.

\subsection{Dataset}
\label{sec:data}
Channel impulse response (CIR) is a representation of convolution between transmitter and receiver. Typically, a CIR represents the signal properties such as reflection, absorption, scattering, and diffraction~\cite{altstidl2021accuracy}. The features used in this work are extracted from raw CIRs. These features are CIR Area (overall CIR energy), CIR total power, CIR First Path / strongest path amplitude ratio, CIR fist path / strongest path time difference, CIR Spectral power, first path width (at half prominence), first path prominence and distance. These 8 features represent a single CIR. The setup has 6 sensors in total. So there are all together 48 features representing a class. There are 6 classes all together namely \emph{left}, \emph{back}, \emph{front}, \emph{driver seat}, \emph{back seat}, \emph{right}. A total of 600 samples are collected from diversified environments and surroundings. Among these 300 samples were used for training and 300 samples for testing. 

\subsection{Network Architecture}
\label{sec:arch}

We evaluated several neural networks such as multi layer perceptron (MLP) network, radial basis functions (RBF) and its variants on the above dataset. Limited by the specific structure of the dataset we selected only variants of fully connected layers. 

\textbf{MLP}: The MLP architecture is a two 32 neurons dense layers which are sandwiched between the two batch normalization layers. The final classification layer is a softmax layer that determines the class to which an input sample belongs. 

\textbf{Single RBF}: A single RBF layer is sandwiched between two batch normalization layers is used as another model. The classification layer is a standard softmax layer for the single layer RBF network as well. The general RBF neuron is represented as shown below:

\begin{equation}
  \phi \left ( x \right ) = e^{-\gamma \left \| x-c_{i} \right \|}
  \label{eq:rbf}
\end{equation}

where, $\gamma$ is $\frac{1}{2\pi\sigma^{2} }$, $c_{i}$ is the centroid. In this case, $\gamma$ and $c_{i}$ are learned through backpropagation. 

\textbf{Multi RBF}:Since the single RBF showed encouraging results several variants of RBF networks are realized. One variant is with multiple $\gamma$s. In this version $n \in [0.1, \dots, 0.3]$. In this work we mention this network as multi-RBF. 

\begin{equation}
  \phi \left ( x \right ) = \sum_{0.1}^{0.3}e^{-\gamma_{n} \left \| x-c_{i} \right \|}
  \label{eq:multirbf}
\end{equation}

\textbf{Multi RBF NR L4}: Another variant is the noise regularized multi RBF (named as RBF NR), L4 norm-based RBF (named as RBF L4) and another variant is noise regularized RBF L4. We present the results for these models in Section 4. The output of an RBF L4 is represented as:

\begin{equation}
  \phi \left ( x \right ) = \sum_{0.1}^{0.3}e^{-\gamma_{n} \left \| x-c_{i} \right \|_{4}}
  \label{eq:multirbf4}
\end{equation}

where, $x$ is the input, $\gamma$ is $\frac{1}{2\pi\sigma^{2} }$, $c_{i}$ is the centroid and $\left\| \cdot \right\|_{4}$ is the L4 distance between the input and the centroids. Here, $n \in [0.1, \dots, 0.3]$ and $c_{i}$ are learned through the backpropagation.

\textbf{Multi RBF NR L4 MH}: Based on the performance comparison on all the stated networks we proposed a multi-head self-supervised architecture where one head performs the denoising reconstruction of the input signal and another head performs classification. This way the RBF layers learned the features that are noise invariant and class specific features. Detailed architectural graph of the network is illustrated in the \cref{fig:selfSupgraph}. 

\textbf{Multi RBF NR L4 MH GC}: The architecture of the network is the same as the Multi RBF NR L4 MH. However we trained the network using the gradient centralization method~\cite{yong2020gradient}. 

\textbf{Multi RBF NR L4 MH AdX}: The architecture of the neural network is the same as Multi RBF NR L4 MH. However we trained the network using the custom regression loss as stated in \cref{eq:Tloss}.

\subsection{Loss Functions}
\label{sec:loss}

We trained all the models with cross entropy loss as a baseline loss functions. The performance of these models are tabulated in \cref{tab:classAcc} except for multi-head neural networks. In a multi-head neural network, one head is trained with a regression loss and another head is a cross entropy loss. Our regression loss is stated in \cref{eq:Tloss}, \cref{eq:mseloss} and \cref{eq:AdXloss}.

\textbf{Regression loss}:

\begin{equation}
  \mathcal{L}_{T_{mse}} = \mathcal{L}_{mse} + \mathcal{L}_{mse_{AdX}} 
    \label{eq:Tloss}
\end{equation}
where, $\mathcal{L}_{T_{mse}}$ is the total loss, $\mathcal{L}_{mse}$ is the mean square error between the true sample and predicted sample and $\mathcal{L}_{mse_{AdX}}$ is the mse between the true sample and the adversarial sample of the $\mathcal{L}_{mse}$.

\begin{equation}
  \mathcal{L}_{mse} = \frac{1}{N}\sum_{i=1}^{N}(x_i-\hat{x_i})^2 
    \label{eq:mseloss}
\end{equation}
where, $N$ is the number of samples, $x_i$ is the true sample, and $\hat{x_i}$ is the predicted approximation of the true sample by the model.
\begin{equation}
  \mathcal{L}_{mse_{AdX}}  = \frac{1}{N}\sum_{i=1}^{N}(x_i-sign(\nabla_{\mathcal{L}_{mse}J(\theta,\mathcal{L}_{mse},y))}))^2 
    \label{eq:AdXloss}
\end{equation}
where, $N$ is the number of samples, $x_i$ is the true sample, $J(\theta,\mathcal{L}_{mse},y)$ is the cost function, $\theta$ is the number of parameters, $\mathcal{L}_{mse}$ is the input to the model, and $y$ is the target.


\subsection{Experiments}
All the networks have the same ADAM optimizer, 0.001 learning rate and 1000 epochs of training. Due to the limited availability of the data samples we performed the experiments in 1-fold, 10-fold and 100-fold accuracy over the dataset.
The expected outcome is that all the data points belonging to a certain class should lie very close to each other (intra class) and at the same time should be very far from the different class data points (inter class). 
Any overlap of the datapoints into a different cluster indicates that the features are very similar, and the classifier is unable to clearly distinguish. 
These results are presented in \cref{sec:res}. The networks were subjected to sensitivity analysis. In these sensitivity analysis plots, accuracy vs multiplication factor ($\epsilon$) of the FGSM \cite{goodfellow2014explaining} gradient was considered. The FGSM attack is modeled as stated in \cite{goodfellow2014explaining}. We considered only the basic version of the adversarial attack because the repetitive frequency of the UWB pulses are high and generating other types of adversarial attacks in the real-time scenarios as per the protocol is not feasible \cite{staderini2001everything}, \cite{kolli2022test}.
After the careful analysis of the sensitivity analysis plots, we proposed a multi-head self-supervised multi-RBF neural network whose results outperformed the previous networks. The representational architecture of the proposed network is illustrated in \cref{fig:selfSupgraph}.

\section{Results}
\label{sec:res}
Several dense neural network architectures are used on the privately collected test dataset. To compensate the lower dataset size we performed 1-Fold, 10-Fold, and 100-Fold validation. Each model is trained for the same loss function, learning rate and epochs. The performances of these architectures are tabulated in the \cref{tab:classAcc}. Performance of the networks under adversarial attacks with $\epsilon = 0.1$ are tabulated in \cref{tab:AdXAcc}. It is observed that our network performs better than the rest of the networks in all three types of attacks except for the BIM. In this case it is the second best.
\begin{table}[]
  \footnotesize
  \centering
  \caption{Localization performance of neural networks for the pre-computed features. Best results are in bold.}
  \renewcommand{\arraystretch}{1.2}
    \begin{tabular}{p{2.9cm}ccc}
    
    \toprule
    \multirow{2}{*}{{ Model }} & \multicolumn{3}{c}{{Accuracy}} \\
    \cline{2-4}
    & {1 Fold} & {10 Fold} & {100 Fold} \\
    \hline
    CIR$_{CNN}$\cite{kolli2022test} & - & 0.92 & - \\
    MLP & 0.963 & 0.956 & 0.956  \\ 
    single-RBF & 0.927 & 0.915 & 0.923\\
    single-RBF-MH & 0.913 & 0.907 & 0.920 \\
    multi-RBF & \textbf{0.987} & \textbf{0.984} & \textbf{0.983}\\ 
    multi-RBF-NR-MH & 0.933 & 0.93 & 0.928 \\ 
    multi-RBF-L4-NR & 0.967 & 0.972 & 0.974\\ 
    \bottomrule
  \end{tabular}
  
  \label{tab:classAcc}
\end{table}

\begin{table}[]
  \small
  \centering
  \caption{Performance of neural networks under adversarial attacks ($\epsilon = 0.1$). Best results are in bold and second best are underlined.}
  \renewcommand{\arraystretch}{1.2}
    \begin{tabular}{p{2.9cm}ccc}
    
    \toprule
    \multirow{2}{*}{{ Model }} & \multicolumn{3}{c}{{Accuracy}} \\
    \cline{2-4}
    & {FGSM} & {BIM} & {PGD} \\
    \hline

    MLP & 0.16 & 0.12 & 0.21  \\ 
    single-RBF & 0.013 & 0.01 & 0.10\\
    multi-RBF & 0.03 & 0.01 & 0.08\\ 
    multi-RBF-L4-NR & 0.08 & 0.02 & 0.04\\ 
    multi-RBF-NR-L4-MH-AdX & \textbf{0.53} &  \underline{0.09} & \textbf{0.59} \\ 
    \bottomrule
  \end{tabular}
  
  \label{tab:AdXAcc}
\end{table}

In the \cref{tab:classAcc} the features selected are outperforming the CNN model trained with the same dataset but with complete CIRs \cite{kolli2022test} alone. We achieved 6\% improvement with our custom features rather than the complete CIRs. Substantial improvement of our model over standard architectures under various types of adversarial attacks are tabulated in \cref{tab:AdXAcc}. 

\begin{figure}[h!]
  \centering
   \includegraphics[width=0.48\textwidth, height=3cm]{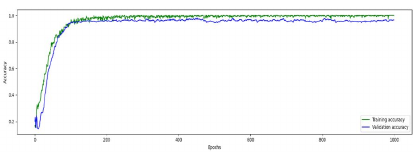}
   \caption{Performance of our model on training and testing dataset. Training accuracy is in green color and testing accuracy is in blue. }
   \label{fig:traintest}
\end{figure}

\begin{figure}[ht!]

\begin{minipage}[b]{1.0\linewidth}
  \centering
  \centerline{\includegraphics[width=1.0\textwidth, height=0.15\textheight]{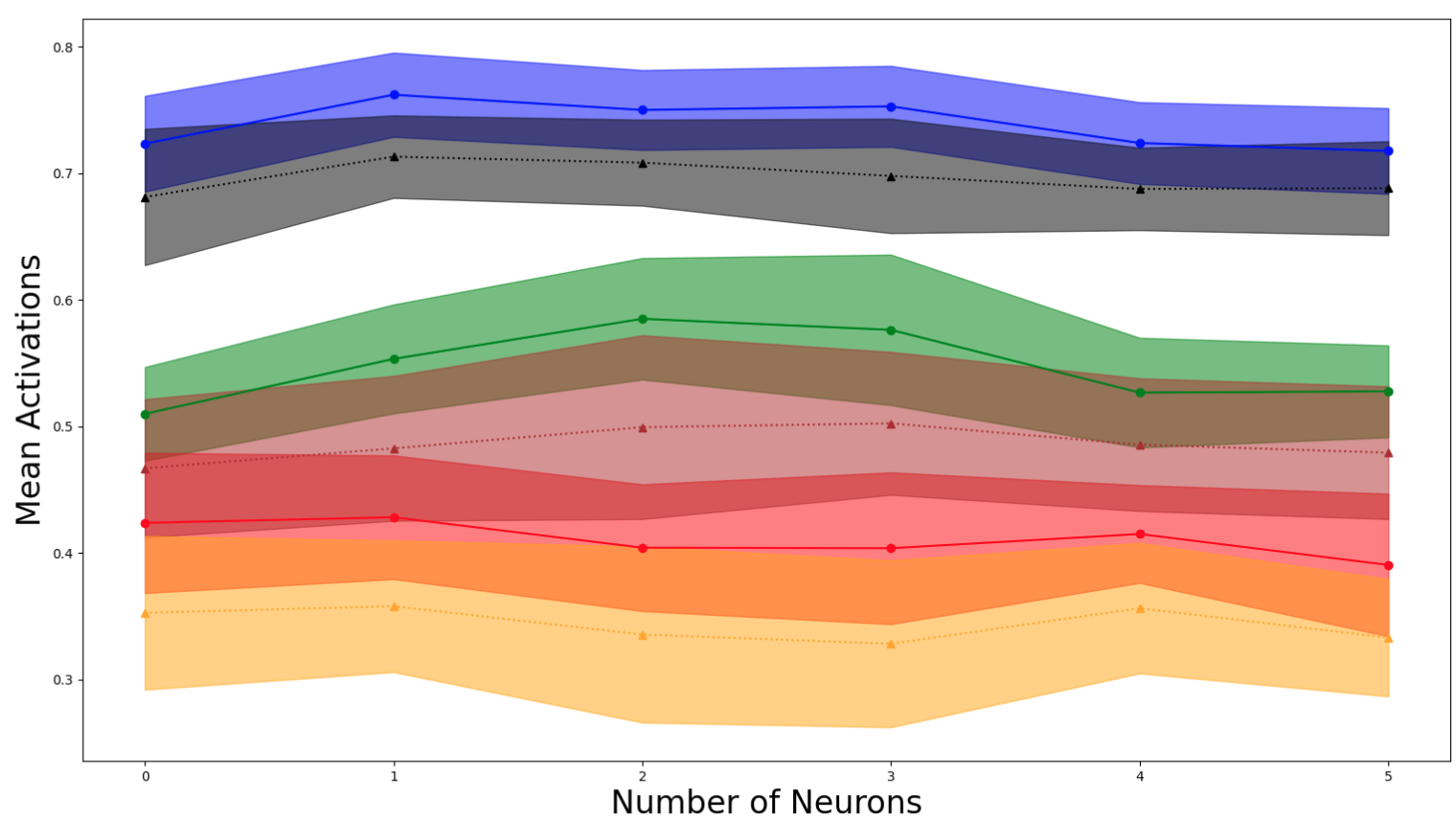}}
  \centerline{(a) Mean activation map of  Multi RBF NR L4}\medskip
\end{minipage}
\begin{minipage}[b]{1.0\linewidth}
  \centering
  \centerline{\includegraphics[width=1.0\textwidth, height=0.15\textheight]{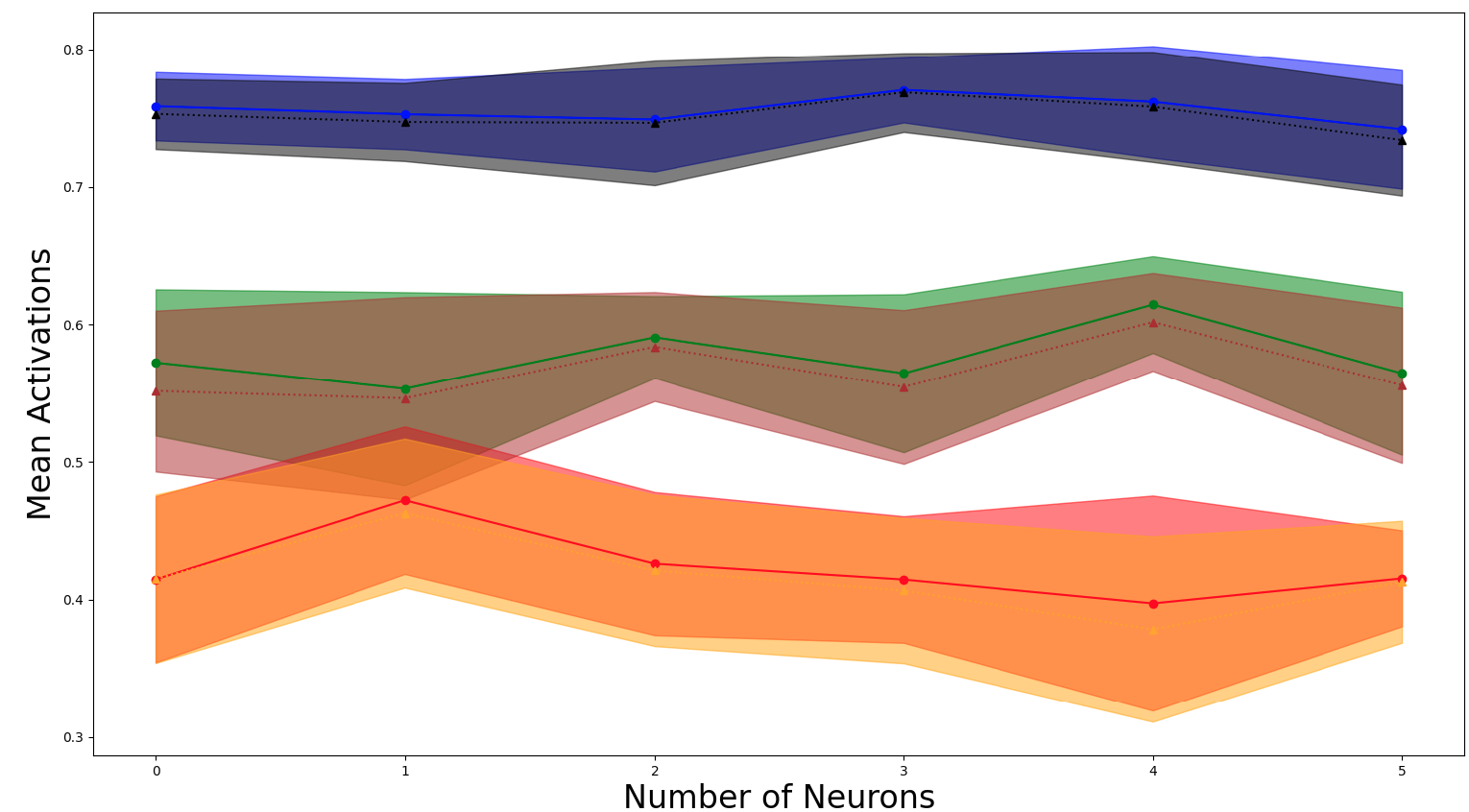}}
  \centerline{(b) Mean activation map of Multi RBF NR L4 MH AdX}\medskip
\end{minipage}
%
\caption{Mean activation maps of RBF neurons of three branches. The blue, green and red colors indicate the mean activations of the RBF layers when clean samples are given. Colors black, brown and yellow indicate mean activations under adversarial sample. }
\label{fig:res}
\end{figure}

\cref{fig:traintest} illustrates the performance of the network at the time of training and testing. This indicates that our model is not overfitting and trained well. The designed self-supervised multi head network also achieved 67\% better adversarial robustness at certain levels without any adversarial training (see \cref{fig:AdRgraph}). \cref{fig:res} illustrates the mean activation maps for the networks Multi RBF NR L4 and  Multi RBF NR L4 MH AdX under clean and adversarial examples. Ideally, the distance between the clean activations should be far from the adversarial examples. 

\section{Conclusion}
Our model trained with noise regularized encoder-decoder and self-supervised model achieved better adversarial robustness over the other baseline models trained in this task. The models realized have achieved 6\% better performance for clean samples, whereas 37\% and 38\% better results for FGSM and PGD attacks over other architectures. We also demonstrated that at certain conditions our self supervised cascaded RBF neuron architecture out performed the standard model by 67\%. The designed models are lighter in size (82.0kB) and can be deployed on to lower memory resources such as micro controller.




\bibliographystyle{IEEEbib}
\bibliography{strings,refs}

\end{document}